\documentclass[10pt,twocolumn,letterpaper]{article}

\usepackage[pagenumbers]{iccv} 

\usepackage{graphicx}
\usepackage{amsmath}
\usepackage{amssymb}
\usepackage{booktabs}
\usepackage{multirow}
\usepackage{cases}

\definecolor{iccvblue}{rgb}{0.21,0.49,0.74}
\usepackage[pagebackref,breaklinks,colorlinks,allcolors=iccvblue]{hyperref}

\def\paperID{****} 
\def\confName{ICCV}
\def\confYear{2025}

\title{Signs as Tokens: A Retrieval-Enhanced Multilingual Sign Language Generator}
\author{Ronglai Zuo, Rolandos Alexandros Potamias, Evangelos Ververas, Jiankang Deng, Stefanos Zafeiriou \\
Imperial College London \\
{\tt\small \href{https://2000zrl.github.io/soke/}{https://2000zrl.github.io/soke/}}
}

\begin{document}
\twocolumn[{
\maketitle
\begin{center}
\vspace{-4mm}
\includegraphics[width=0.95\textwidth]{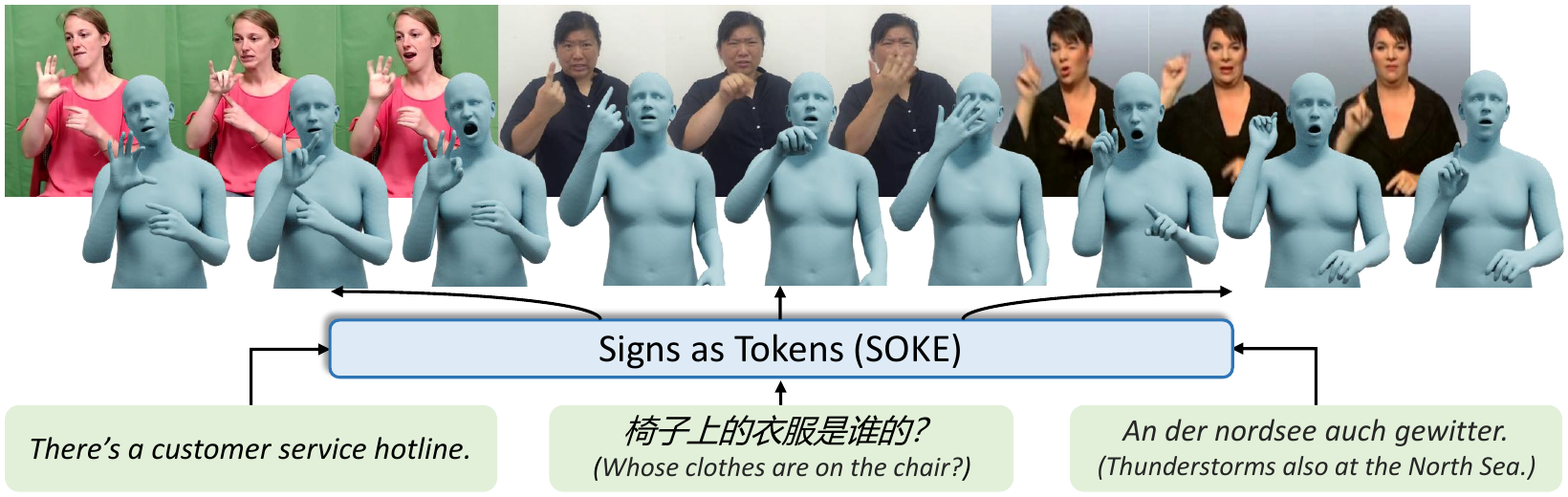}
\vspace{-2mm}
\captionof{figure}{We propose Signs as Tokens (SOKE), a unified sign language generator that can produce multilingual signs from text inputs. (Left: American sign language; Middle: Chinese sign language; Right: German sign language.)}
\label{fig:teaser}
\end{center}
}]

\begin{abstract}
Sign language is a visual language that encompasses all linguistic features of natural languages and serves as the primary communication method for the deaf and hard-of-hearing communities. Although many studies have successfully adapted pretrained language models (LMs) for sign language translation (sign-to-text), the reverse task—sign language generation (text-to-sign)—remains largely unexplored. In this work, we introduce a multilingual sign language model, Signs as Tokens (SOKE), which can generate 3D sign avatars autoregressively from text inputs using a pretrained LM. To align sign language with the LM, we leverage a decoupled tokenizer that discretizes continuous signs into token sequences representing various body parts. During decoding, unlike existing approaches that flatten all part-wise tokens into a single sequence and predict one token at a time, we propose a multi-head decoding method capable of predicting multiple tokens simultaneously. This approach improves inference efficiency while maintaining effective information fusion across different body parts. To further ease the generation process, we propose a retrieval-enhanced SLG approach, which incorporates external sign dictionaries to provide accurate word-level signs as auxiliary conditions, significantly improving the precision of generated signs. Extensive qualitative and quantitative evaluations demonstrate the effectiveness of SOKE.

\end{abstract}
    
\vspace{-5mm}
\section{Introduction}
\label{sec:intro}

Sign language is the primary communication method utilized by the deaf and hard-of-hearing communities, embodying all linguistic properties found in spoken languages, such as discrete semantic units and grammatical structure \cite{yin2021including,gong2024llms}. 
As an attempt to mitigate the barrier between sign and spoken languages, several methods have been developed to tackle sign language processing by unfolding the task into two research problems: sign language translation (SLT, sign-to-text) \cite{hu2023signbert+,yin2022mlslt,zhou2023gloss,wongsign2gpt,stmc_jour,yao2023sign,li2025uni-sign,tan2025multilingual} and sign language generation (SLG, text-to-sign) \cite{saunders2020progressive,baltatzis2024nsa,yu2025signavatars,zuo2024simple,shi2024pose}.

In contrast with SLT, the usage of language models (LMs) in SLG remains relatively unexplored. 
Most existing methods approach SLG as a visual content generation task (\eg, video, keypoint, or motion), using GANs~\cite{shi2024pose,saunders2022signing} or diffusion models~\cite{baltatzis2024nsa,signgen,fang2023signdiff,tang2025discrete,tang2025sign}. However, these works neglect the linguistic nature of sign languages, leading to suboptimal performance and missing benefits of pretrained LMs such as generalizability and scalability \cite{tian2024var,kaplan2020scaling,henighan2020scaling}.
In particular, sign languages share the fundamental linguistic properties and the discrete structure of spoken languages \cite{yin2021including,gong2024llms}, which necessitates processing using autoregressive language modeling techniques \cite{yin-etal-2024-t2s}.

To formulate SLG in the framework of language modeling, a crucial step is mapping signs into discrete tokens. A common approach to discretizing sign languages involves using glosses—the written form of signs—as an intermediate representation \cite{zuo2024simple,saunders2022signing,huang2022dualsign,fang2024signllm}. However, glosses require extensive annotation efforts and impose a predefined information bottleneck, which fail to fully capture the rich semantics of sign languages \cite{muller2022considerations}.
To address these limitations and better capture the multi-cue property of sign languages \cite{zuo2022c2slr,cosign,stmc}, we leverage a vector quantized-variational auto-encoder (VQ-VAE)-based tokenizer \cite{yi2023generating,lu2024humantomato,xie2024g2p} that learns mappings between continuous sign motions and discrete tokens for each body part (upper body, left hand, and right hand). 
During decoding, existing autoregressive motion generation methods \cite{yi2023generating,lu2024humantomato} typically flatten tokens from different body parts into a single sequence, resulting in inefficient inference due to tripled decoding steps. To overcome this challenge, we propose a novel multi-head decoding method capable of predicting multiple tokens at a time while effectively fusing information from different body parts in real time.
Furthermore, given the limited vocabulary size of continuous sign language datasets, we enhance the model's generalization by instantiating the LM as a multilingual LM \cite{mbart}, enabling it to handle a broader and more diverse domain of discourse. To support multilingual SLG in a unified framework, we train the proposed model on a curated multilingual sign language dataset encompassing American, Chinese, and German sign languages \cite{duarte2021how2sign,csl-daily,2014T}.

Recently, retrieval-augmented generation (RAG) \cite{rag_benchmark,rag_survey,wang2024searching} has proven effective in enhancing the interpretability and accuracy of content generated by large language models. Similarly, several SLG methods \cite{zuo2024simple,saunders2022signing} leverage glosses to retrieve sign dictionaries and directly incorporate retrieved signs into the outputs, improving the interpretability of generated signs. However, beyond the inherent limitations of glosses, these methods often face challenges in handling co-articulations between adjacent signs. This is because dictionary signs are typically recorded at the word level, whereas SLG models operate at the sentence level, leading to potential unnaturalness in the generated sequences.
To address these issues, we propose a retrieval-enhanced SLG approach that utilizes motion tokens from retrieved dictionary signs as additional conditions for language modeling. This approach avoids the unnaturalness of directly copying signs into model outputs while still benefiting from the precision of dictionary signs.
To sum up, the contributions of this paper can be summarized as:
\begin{itemize}
    \item We propose a novel SLG approach, Signs as Tokens (SOKE). With the aid of a pretrained LM, SOKE can handle multiple sign languages in a unified model.
    \item We propose a multi-head decoding method that reduces inference latency and efficiently fuses information from different body parts, enabling seamless integration of a decoupled tokenizer with the LM.
    \item By leveraging external sign dictionaries, we introduce a retrieval-enhanced SLG approach to improve the precision of generated signs.
    \item We curate two sign language datasets \cite{csl-daily,2014T} with accurate SMPL-X \cite{smplx} poses to facilitate multilingual SLG research. Experiments demonstrate that our SOKE achieves state-of-the-art performance on three challenging benchmarks, How2Sign \cite{duarte2021how2sign}, CSL-Daily \cite{csl-daily}, and Phoenix-2014T \cite{2014T}, using a single unified model.
\end{itemize}

\begin{figure*}[t]
\centering
\includegraphics[width=1.0\linewidth]{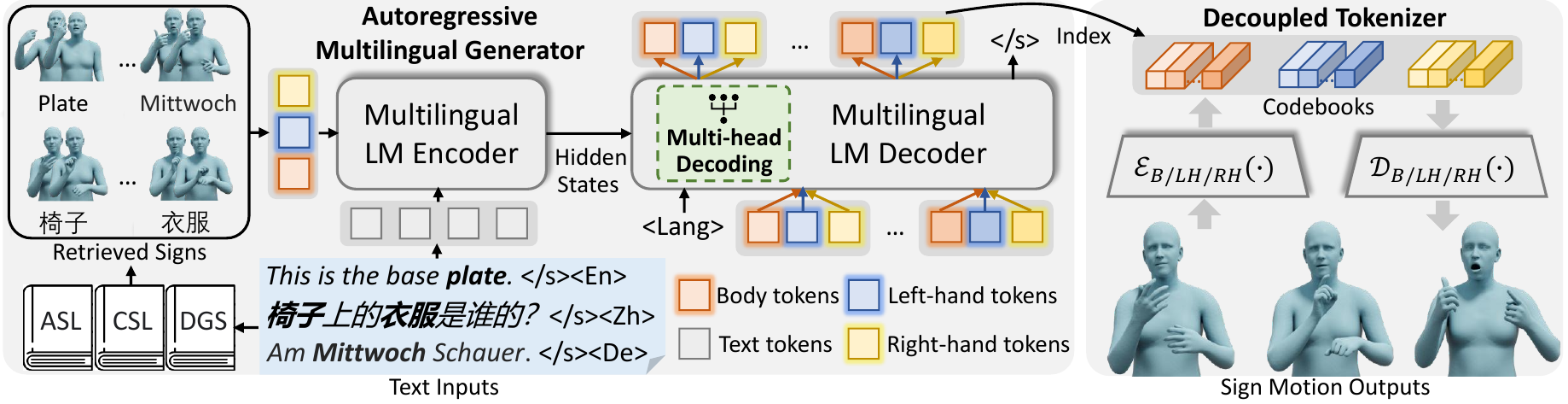}
\vspace{-6mm}
\caption{An overview of our proposed method, Signs as Tokens (SOKE). We begin by training a VQ-VAE-based decoupled tokenizer to map continuous sign motions into discrete tokens for various body parts (upper body, left hand, and right hand). These sign motion tokens are then integrated into the text vocabulary of a pretrained language model, which serves as the backbone of our autoregressive multilingual generator (AMG). Given a text input, the AMG first retrieves word-level signs from external dictionaries, appends their motion tokens to the text tokens, and feeds them into the language model encoder. During decoding, our novel multi-head decoding strategy generates motion tokens for all body parts simultaneously at each time step. Finally, the derived motion tokens are used to reconstruct sign avatars.}
\vspace{-2mm}
\label{fig:framework}
\end{figure*}
\section{Related Work}
\label{sec:rel}

\noindent\textbf{Sign Language Generation.} 
Early gloss-based methods \cite{zuo2024simple, saunders2022signing, zelinka2020neural, xie2024g2p, huang2021towards, tang2022gloss, huang2022dualsign} approach SLG under a text-to-gloss-to-sign framework. 
Several works \cite{arkushin2023ham2pose,saunders2020progressive,saunders2021continuous} propose to directly regress 2D joints to represent sign poses, simplifying the modeling process, however, they suffer from mode collapse due to the limited scale of data.
Recent state-of-the-art SLG works can be categorized into two classes: the first group of methods \cite{baltatzis2024nsa,fang2023signdiff} employs diffusion models to generate sign motions conditioned on text inputs; the second group of methods \cite{fg2024signavatar,yu2025signavatars,yin-etal-2024-t2s} adopts a tokenizer-LM two-stage autoregressive generation approach. 
Among these, \cite{yu2025signavatars} incorporates costly gloss and HamNoSys notations \cite{arkushin2023ham2pose} to train a semantic adapter, while \cite{fg2024signavatar} focuses on generating signs from a single word, limiting their applicability.
Moreover, a notable gap exists in the literature regarding unified models for multilingual SLG. In this work, we address these gaps by utilizing a decoupled tokenizer and a pretrained multilingual LM, integrated with our customized multi-head decoding method and retrieval-enhanced approach, to enable text-driven multilingual SLG.

\noindent\textbf{Text-to-Motion Generation.} 
Text-to-motion generation (T2MG) is an important research area with potential applications in human-computer interaction and robotics. 
Recent T2MG methods can be categorized into two main types: 1) Diffusion-based approaches \cite{zhang2024motiondiffuse,shafirhuman,chen2023executing}, which utilize latent diffusion \cite{latentdiff} to generate motions conditioned on text inputs. 2) Autoregressive approaches \cite{jiang2023motiongpt,aaai_motiongpt,zhou2024avatargpt,t2m-gpt,guo2022tm2t,lu2024humantomato}, which draw inspiration from text generation methods by first tokenizing motion sequences and then using LMs to perform autoregressive generation.
When decoding with a decoupled tokenizer \cite{lu2024humantomato,yi2023generating}, these autoregressive approaches simply flatten tokens from various body parts into one sequence, causing addtional decoding steps. In this work, we propose a novel multi-head decoding method that can efficiently predict motion tokens for all body parts at one single step, significantly reducing inference latency.

\noindent\textbf{Retrieval-Enhanced Generation.}
Using the concept of retrieval to enhance generation models has proven effective across various domains \cite{rag_benchmark,rag_survey,wang2024searching,Golatkar2024cpr,Xu2024rag_ego,ding2024realgen}. As a pioneering effort, Piktus \etal \cite{lewis2020retrieval} propose to retrieve relevant documents before generating answers for knowledge-intensive NLP tasks. FLARE \cite{jiang2023active} introduces selective retrieval, activating it only when low-confidence tokens appear in the generation process. ToG \cite{sunthink} leverages retrieved knowledge to enhance the reasoning capabilities of large language models. In the field of image generation, retrieval has also been utilized to enable diffusion models to generate out-of-distribution images \cite{sheyninknn}.
In the context of SLG, several methods \cite{zuo2024simple,saunders2022signing} adopt gloss-based retrieval to obtain dictionary signs and approach SLG in a retrieve-and-concatenate manner. However, glosses have inherent limitations, and such methods often suffer from unrealistic transitions between signs. In this paper, we propose a simple yet effective retrieval-enhanced SLG approach that uses word-level signs as auxiliary conditions, addressing these limitations and improving the overall quality of generations.
\section{Methodology}
\begin{figure*}[t]
\centering
\includegraphics[width=1.0\linewidth]{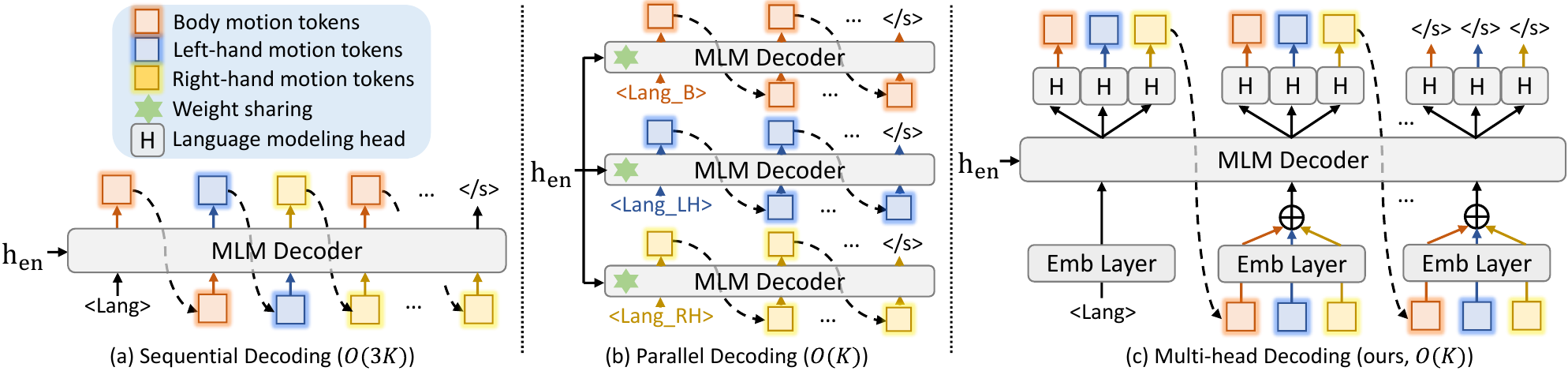}
\vspace{-5mm}
\caption{Comparison of various decoding methods. (a) Sequential decoding flattens all motion tokens into a single sequence, requiring an inefficient $3K$ decoding steps, where $K$ is the target motion sequence length. (b) Parallel decoding improves efficiency by decoding body parts in parallel but lacks information fusion, which may lead to suboptimal performance. (c) Our proposed multi-head decoding balances efficiency and fusion by predicting three tokens simultaneously via different heads, with input embeddings as a weighted average of part-wise token embeddings.}
\vspace{-3mm}
\label{fig:decoding}
\end{figure*}

An overview of our method is shown in Figure \ref{fig:framework}. Our approach mainly consists of two parts: a decoupled tokenizer (DETO) and an autoregressive multilingual generator (AMG). In the first stage, DETO learns mappings between continuous sign motions and discrete tokens for each body part (Section \ref{sec:tokenizer}). In the second stage, AMG is trained to autoregressively generate motion tokens with the proposed multi-head decoding strategy (Section \ref{sec:gen}) conditioned on text inputs and retrieved dictionary signs (Section \ref{sec:retrieval}).

\subsection{Data Preparation}
\label{sec:data}
Although existing SLG works \cite{baltatzis2024nsa,yu2025signavatars} have made efforts in curating high-quality SMPL-X poses for an American sign language (ASL) dataset, How2Sign \cite{duarte2021how2sign}, there is a notable lack of 3D annotations for other continuous sign language datasets, impeding multilingual SLG research.
In this work, we rely on a two-stage approach, leveraging state-of-the-art 3D pose estimation methods \cite{lin2023osx,potamias2024wilor}, to extract high-fidelity SMPL-X poses for CSL-Daily \cite{csl-daily} and Phoenix-2014T \cite{2014T}, two widely-adopted datasets of Chinese and German sign languages (CSL and DGS), respectively. 
Specifically, we initially estimate a set of rough 3D body poses for each signer using OSX \cite{lin2023osx} along with accurate 3D hand poses from WiLoR~\cite{potamias2024wilor}. 
Given that OSX often falls short in accurately capturing the arm poses, we follow~\cite{baltatzis2024nsa} and refine the upper body joint rotations by minimizing a re-projection loss between the estimated joints and the detected 2D joint keypoints from Mediapipe~\cite{mediapipe}. 
Using this process we can ensure accurate whole-body annotations. 
We represent each sign motion sequence as $\mathbf{S} \in \mathbb{R}^{T\times d}$, where $T$ is the sequence length and $d=133$ denotes the number of SMPL-X parameters, including 11 upper body joints, 30 hand joints, and 10 expression parameters. Please check the supplementary materials for more details and quantitative evaluations of our pose fitting pipeline.

\subsection{Decoupled Tokenizer}
\label{sec:tokenizer}
Sign languages exhibit a multi-cue property \cite{stmc,zuo2022improving,cosign,best,zhao2024self,hu2021hand}, where the semantics of a sign are conveyed simultaneously through body and hand movements. Inspired by this, we leverage a decoupled tokenizer \cite{lu2024humantomato,yi2023generating,xie2024g2p}, comprising three VQ-VAEs \cite{vqvae}, to independently model key regions: the upper body and both hands. An illustration is available in the supplementary materials.

Given a $T$-frame sign motion input $\mathbf{S} \in \mathbb{R}^{T\times d}$, we first decompose it into three part-wise motion sequences based on the kinematic tree of SMPL-X: $\mathbf{S}^{p} \in \mathbb{R}^{T\times d_p}$, where $p\in \{B, LH, RH\}$. Moreover, we build three distinct VQ-VAEs, where each of them consists of an encoder $\mathcal{E}_p(\cdot)$, a decoder $\mathcal{D}_p(\cdot)$, and a learnable codebook $\mathbf{Z}^p \in \mathbb{R}^{N_Z^p\times C}$, where $N_Z^p$ represents the number of codes and $C$ denotes the code dimension. 
For each motion sequence, the corresponding encoder first projects the sequence into a latent space: $\mathbf{S}_f^p = \mathcal{E}_p(\mathbf{S}^p) = \{s_{f,i}^p\}_{i=1}^{T_f} \in \mathbb{R}^{T_f\times C}$, using a stack of 1D-CNN layers. 
Then, for each pose we can derive a set of discrete tokens $\mathbf{\hat{Z}}^p = \{\hat{z}_i^p\}_{i=1}^{T_f}$ using a quantizer $Q(\cdot)$, which searches the nearest neighbor from the codebook $\mathbf{Z}^p$:
\begin{equation}
    \hat{z}_i^p = Q(s_{f,i}^p) = \arg\min_{z_j\in\mathbf{Z}^p} \|s_{f,i}^p - z_j\|_2, \ j\in [1, N_Z^p].
\end{equation}

We then feed the obtained token sequence to the corresponding part-decoder $\mathcal{D}_p$ to reconstruct the input motions: $\mathbf{\hat{S}}^p = \mathcal{D}_p(\mathbf{Z}^p)$. Following \cite{jiang2023motiongpt}, the objective function of each VQ-VAE is defined as: $\mathcal{L}_{vq}^p = \mathcal{L}_{rec}^p + \mathcal{L}_{emb}^p + \mathcal{L}_{com}^p$, where $\mathcal{L}_{rec}^p = \|\mathbf{\hat{S}}^p - \mathbf{S}^p\|_2^2$ denotes the reconstruction loss,  $\mathcal{L}_{emb}^p$ is the embedding loss, and $\mathcal{L}_{com}^p$ is the commitment loss, respectively.

\subsection{Autoregressive Multilingual Generator}
\label{sec:gen}
Using obtained discrete sign representations, we can respect the discrete properties of sign languages and approach SLG similarly to text generation with a pretrained LM \cite{mbart}. 

\noindent\textbf{Integrated Vocabulary.}
We first gather the discrete tokens for different body parts to construct a part-wise motion vocabulary: $\mathcal{V}_m^p = \{\textless \mathtt{p\_i} \textgreater\}_{i=1}^{N_Z^p}$, \eg, $\textless \mathtt{B\_1} \textgreater$ and $\textless \mathtt{RH\_2} \textgreater$. The overall motion vocabulary is a combination of three part-wise vocabularies: $\mathcal{V}_m = \{\mathcal{V}_m^{B}, \mathcal{V}_m^{LH}, \mathcal{V}_m^{RH}\}$.
Following \cite{mbart}, we further define a series of language identifier tokens: $\mathcal{V}_l = \{\textless\mathtt{ASL}\textgreater, \textless\mathtt{CSL}\textgreater, \textless\mathtt{DGS}\textgreater\}$, to prompt the LM with the information of target sign language. The integrated vocabulary for the generator is defined as: $\mathcal{V} = \{\mathcal{V}_t, \mathcal{V}_m, \mathcal{V}_l\}$, which lays a foundation for the subsequent LM fine-tuning.

\noindent\textbf{Sequential Decoding.}
Following previous motion generation works \cite{lu2024humantomato,yi2023generating}, a naive decoding method is simply flattening tokens from each body part into a single sequence with tripled length: $\mathbf{Y} = \{y_1^{B}, y_1^{LH}, y_1^{RH}, \dots, y_K^{B}, y_K^{LH}, y_K^{RH}\}$, where $y_k^p \in \mathcal{V}_m^p$ denotes the part-wise motion token, and $K$ denotes the length of the target sign motion sequence. The decoding process can then be formulated as:
\vspace{-5mm}
\begin{align}
\label{eq:seq}
    &P(\mathbf{Y}|\mathbf{h}_{en}) = \prod_{k=1}^K P(y_k^{B,LH,RH} | y_{<k}^{B,LH,RH}) \\
    &=\prod_{k=1}^K P(y_k^{B} | y_{<k}^{\{p\}}) P(y_k^{LH} |y_k^{B}, y_{<k}^{\{p\}}) P(y_k^{RH} |y_k^{B,LH}, y_{<k}^{\{p\}}), \nonumber
\end{align}
where $\mathbf{h}_{en}$ denotes the last encoder hidden states. For simplicity, we write $y_{<k}^{B,LH,RH}$ as $y_{<k}^{\{p\}}$ and omit $\mathbf{h}_{en}$.
We term this decoding method as sequential decoding (Figure \ref{fig:decoding}(a)), which results in inefficient inference process requiring $3K$ decoding steps.

\noindent\textbf{Parallel Decoding.}
In real-world applications, the model's real-time capability is important. To accelerate the decoding process, a straightforward way is to decompose the target sequence $\mathbf{Y}$ into three part-wise sequences: $\mathbf{Y}^p=\{y_1^p, \dots, y_K^p\}$.
Moreover, three LM decoders are instantiated with shared weights, each one in charge of a single body part. To prompt the decoders with the information of body parts, we replace the language identifier tokens with a new set of special tokens: $\mathcal{V}_l = \{\textless\mathtt{Lang\_p}\textgreater\}$, where $\mathtt{Lang}$ and $\mathtt{p}$ denote the target language and body part, respectively. 

As shown in Figure \ref{fig:decoding}(b), three parallel decoding processes start from one of the above mentioned special tokens, based on the target language and body part. For example, the starting token would be set to \textless$\mathtt{ASL\_B}$\textgreater when the model is required to generate upper-body motions for ASL. 
The decoding process can then be formulated as:
\begin{align}
\label{eq:parallel}
    &P(\mathbf{Y}|\mathbf{h}_{en}) = P(\mathbf{Y}^B) P(\mathbf{Y}^{LH}) P(\mathbf{Y}^{RH}) \\
    &=\prod_{k=1}^K P(y_k^{B} | y_{<k}^{B}) \prod_{k=1}^K P(y_k^{LH} | y_{<k}^{LH}) \prod_{k=1}^K P(y_k^{RH} | y_{<k}^{RH}). \nonumber
\end{align}

Comparing Eq. \ref{eq:parallel} with Eq. \ref{eq:seq}, the parallel decoding strategy significantly reduces the dependence on previous tokens, as the decoding processes for each body part are completely separated. While this formulation enables efficient parallel execution, the overly independent assumption may result in suboptimal performance.

\noindent\textbf{Multi-Head Decoding.}
To balance efficiency and information fusion across body parts, we propose a multi-head decoding strategy. As shown in Figure \ref{fig:decoding}(c), we design three language modeling heads, implemented as fully connected layers, to predict motion tokens for each body part simultaneously at each step. Moreover, the decoder inputs at each step are modified to a weighted average of the token embeddings from different body parts. Specifically, denoting the token embeddings from upper body, left hand, and right hand as $\mathbf{E}^B$, $\mathbf{E}^{LH}$, and $\mathbf{E}^{RH}$, respectively, the input embeddings are defined as: $\mathbf{E}=(1-2\lambda)\mathbf{E}^B + \lambda\mathbf{E}^{LH} + \lambda\mathbf{E}^{RH}$, where $\lambda\in(0,0.5)$ is a hyper-parameter to control the weight of hand embeddings.
The decoding process can then be formulated as:
\begin{align}
\label{eq:mhead}
    &P(\mathbf{Y}|\mathbf{h}_{en}) = \prod_{k=1}^K P(y_k^{B,LH,RH} | y_{<k}^{B,LH,RH}) \\
    &=\prod_{k=1}^K P(y_k^{B} | y_{<k}^{\{p\}}) P(y_k^{LH} | y_{<k}^{\{p\}}) P(y_k^{RH} | y_{<k}^{\{p\}}), \nonumber
\end{align}
which can be viewed as relaxing Eq. \ref{eq:seq} under a conditional independence assumption.

\noindent\textbf{Training and Inference.}
The LM is trained with a standard cross-entropy loss: $\mathcal{L}_{LM}=-\log P(\mathbf{Y}|\mathbf{h_{en}})$.
During inference, we adopt a simple greedy decoding algorithm that the LM always outputs the token with the highest probability at each step. For parallel and multi-head decoding, the inference process terminates as soon as any head or decoder predicts an end-of-sentence token. The obtained token sequences will be fed into the corresponding part-decoder, $\mathcal{D}_p(\cdot)$, to reconstruct sign motions.

\subsection{Retrieval-Enhanced SLG}
\label{sec:retrieval}
\begin{figure}[t]
\centering
\includegraphics[width=1.0\linewidth]{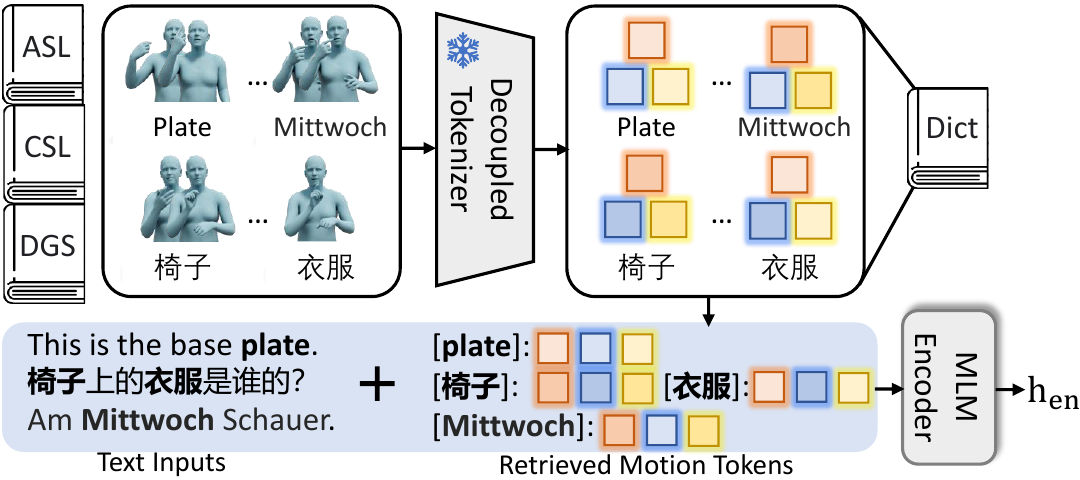}
\vspace{-6mm}
\caption{Illustration of retrieval-enhanced SLG. We utilize external sign dictionaries to retrieve accurate word-level signs, which serve as additional conditions to enhance the generation quality.}
\vspace{-2mm}
\label{fig:ret}
\end{figure}

Inspired by the effectiveness of retrieval-augmented generation in NLP \cite{rag_benchmark,rag_survey,wang2024searching}, we propose to leverage external sign dictionaries to improve the fidelity of generated signs. 

As shown in Figure \ref{fig:ret}, we first build a word-level sign dictionary for each target sign language by utilizing resources from isolated sign language recognition datasets \cite{joze2019ms, li2020word, zuo2023natural, zhao2024masa, dw-dgs} and online sources. Through our pose fitting pipeline, each RGB video in these dictionaries is converted into SMPL-X poses. Next, we employ a well-trained decoupled tokenizer to map the poses into discrete tokens. The resulting dictionary is represented as a set of quadruples $\{(w, m^B, m^{LH}, m^{RH})\}$, where $w$ denotes lemmatized words in the written language, and $m^p$ denotes motion tokens.
If a word has multiple sign instances in the dictionary, we retain only the instance with the lowest reconstruction error after passing through the tokenizer. 

Given a text input $\mathbf{X}=\{x_i\}_{i=1}^L$, consisting of $L$ words\footnote{Since Chinese sentences lack spaces between words, we use Jieba \cite{jieba,niu2023hksl} for word segmentation.}, we collect all words and corresponding tokens found in the dictionary. The final prompt is created by concatenating raw texts $\mathbf{X}$ with the retrieved motion tokens $\{m_j^p\}$, which is then fed into the LM encoder to generate hidden states $\mathbf{h}_{en}$. Several examples are shown in Figure \ref{fig:ret}.
\begin{figure*}[t]
    \centering
    \begin{subfigure}[b]{1.0\textwidth}
        \centering
        \includegraphics[width=\textwidth]{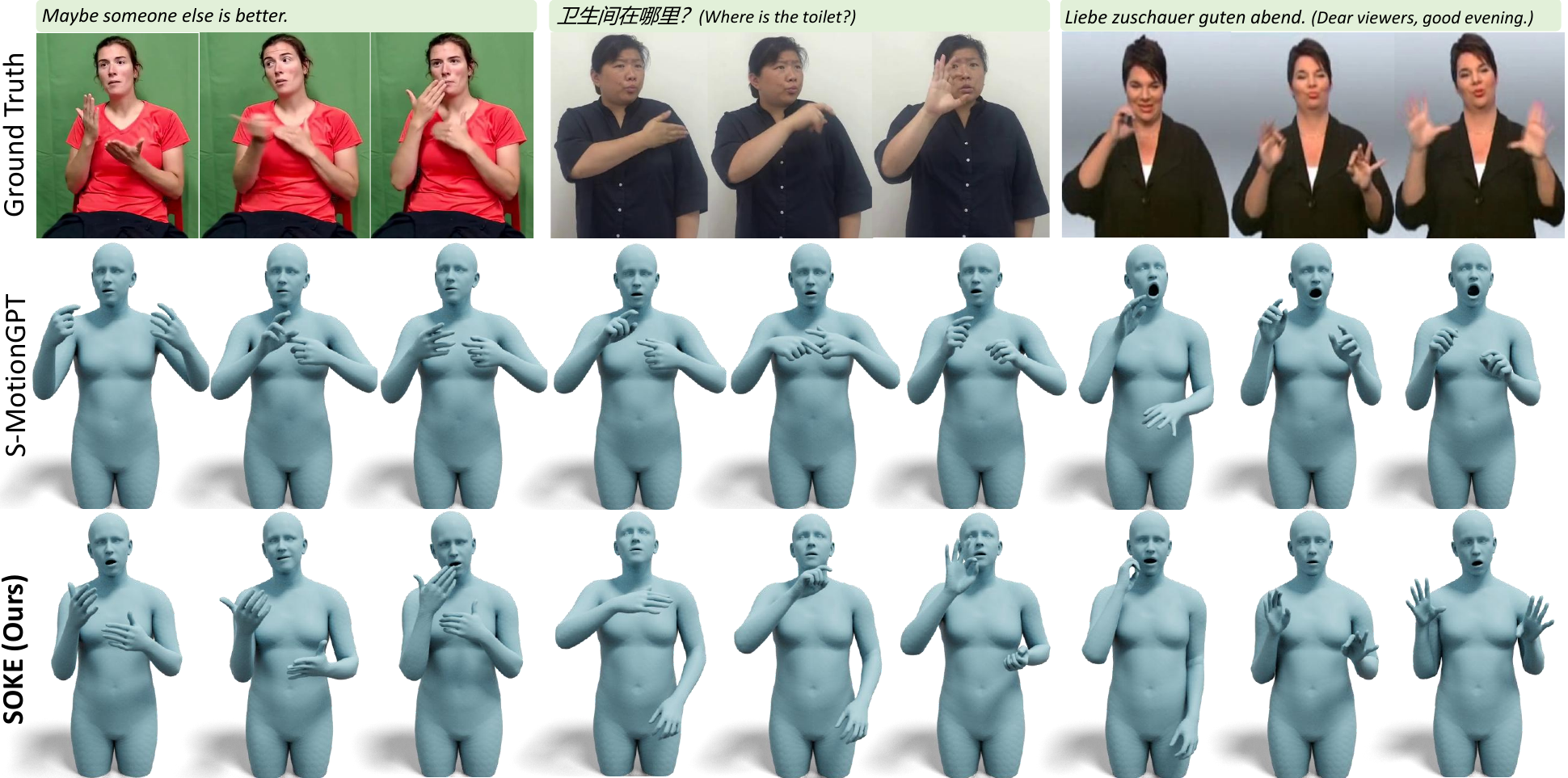}
        \vspace{-5mm}
        \label{fig:vis_1}
    \end{subfigure}
    \hfill
    \begin{subfigure}[b]{1.0\textwidth}
         \centering
         \includegraphics[width=\textwidth]{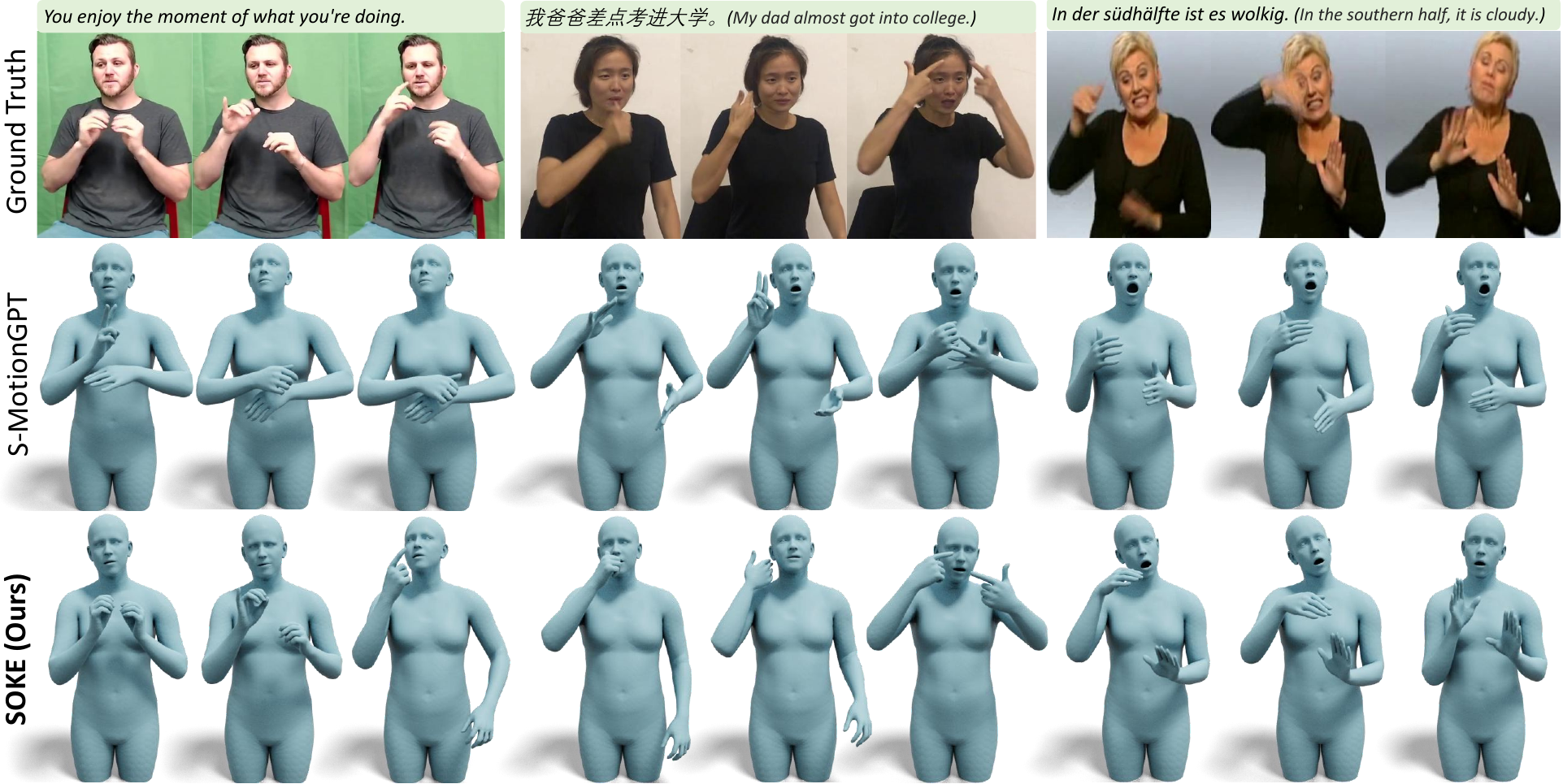}
         \vspace{-6mm}
         \label{fig:vis_2}
     \end{subfigure}
     \hfill

    \vspace{-5mm}
    \caption{Qualitative comparisons of generated signs between our proposed method, SOKE, with the SOTA method, S-MotionGPT \cite{jiang2023motiongpt}, on the test sets of How2Sign (left), CSL-Daily (middle), and Phoenix-2014T (right). }
    \label{fig:vis}
    \vspace{-4mm}
\end{figure*}

\section{Experiments}
\noindent\textbf{Datasets.}
We construct a multilingual sign language dataset by merging three widely-adopted datasets: How2Sign \cite{duarte2021how2sign}, CSL-Daily \cite{csl-daily}, and Phoenix-2014T \cite{2014T}. How2Sign is a large-scale ASL dataset, consisting of 35K video-text pairs. CSL-Daily is the largest existing continuous CSL dataset, containing 20K samples. Phoenix-2014T is a DGS dataset with 8K samples. 
To train SOKE, we use the SMPL-X poses of How2Sign dataset as provided in \cite{baltatzis2024nsa} along with the curated poses of CSL-Daily and Phoenix-2014T.

\noindent\textbf{Evaluation Metrics.}
Since the length of generated signs may differ from the ground truth, we employ the well-established dynamic time warping (DTW) \cite{baltatzis2024nsa,arkushin2023ham2pose} on both procrustes-aligned and original joint position errors (JPE) to measure sequence-level distances between the generated signs and ground truth. Following \cite{baltatzis2024nsa}, we also report back-translation (B-T) scores in BLEU-4 \cite{papineni2002bleu} to assess the interpretability of the generated signs.

\noindent\textbf{Implementation Details.}
For DETO, we empirically set the code numbers to $N_Z^B=96$ for the body and $N_Z^{LH} = N_Z^{RH}=192$ for the left and right hands, respectively, with a code dimension of $C=512$. 
We train DETO on the combined multilingual sign language dataset and sign dictionaries with a batch size of 320 per GPU for 500 epochs, using the AdamW optimizer \cite{adamw} and a cosine learning rate scheduler starting at 2e-4. 
For the LM, we use mBART-large-cc25 \cite{mbart}, which features 12 encoder-decoder layers with a model dimension of 1024 and is widely used in various sign language translation methods \cite{chentwo, Wei_2023_ICCV, chen2022simple, zuo2024towards, jiao2024visual}. We empirically set $\lambda=1/3$ to balance part-wise token embeddings. The LM is fine-tuned with a batch size of 32 per GPU on the multilingual sign language dataset for 150 epochs, employing the same optimizer settings as DETO. All models are trained using 6 RTX 3090 GPUs.

\begin{table*}[t]
\small
\setlength\tabcolsep{3pt}
    \centering
    \resizebox{\linewidth}{!}{
    \begin{tabular}{l|c|ccccc|ccccc|ccccc}
    \toprule
    \multirow{3}{*}{Method} & \multirow{3}{*}{\shortstack{Multi-\\lingual}} & \multicolumn{5}{c|}{How2Sign} & \multicolumn{5}{c|}{CSL-Daily} & \multicolumn{5}{c}{Phoenix-2014T} \\
    & & \multicolumn{2}{c}{DTW-PA-JPE$\downarrow$} & \multicolumn{2}{c}{DTW-JPE$\downarrow$} & B-T$\uparrow$ & \multicolumn{2}{c}{DTW-PA-JPE$\downarrow$} & \multicolumn{2}{c}{DTW-JPE$\downarrow$} & B-T$\uparrow$ & \multicolumn{2}{c}{DTW-PA-JPE$\downarrow$} & \multicolumn{2}{c}{DTW-JPE$\downarrow$} & B-T$\uparrow$ \\
    \cmidrule(r){3-4}\cmidrule(r){5-6}\cmidrule(r){7-7}\cmidrule(r){8-9}\cmidrule(r){10-11}\cmidrule(r){12-12}\cmidrule(r){13-14}\cmidrule(r){15-16}\cmidrule(r){17-17}
    & & Body & Hand & Body & Hand & BLEU-4 & Body & Hand & Body & Hand & BLEU-4 & Body & Hand & Body & Hand & BLEU-4 \\
    
    \midrule
    NAR \cite{hwang2021non} & \texttimes & 13.94 & 11.80 & -- & -- & 5.75 & -- & -- & -- & -- & -- & -- & -- & -- & -- & --  \\
    Prog. Trans.$^*$ \cite{saunders2020progressive} & \texttimes & 14.15 & 11.57 & 14.74 & 30.17 & 2.75 & 15.98 & 12.91 & 16.30 & 32.63 & 3.07 & 13.67 & 11.95 & 15.01 & 31.77 & 4.94 \\
    Text2Mesh$^*$ \cite{stoll2022there}& \texttimes & 13.99 & 13.47 & 15.50 & 32.97 & 7.51 & 13.47 & 12.10 & 13.76 & 30.37 & 5.11 & 13.48 & 12.06 & 14.04 & 31.64 & 5.81 \\
    Adv. Train. \cite{saundersadversarial} & \texttimes & 13.78 & 11.17 & -- & -- & 6.21 & -- & -- & -- & -- & -- & -- & -- & -- & -- & --\\
    T2S-GPT$^*$ \cite{yin-etal-2024-t2s} & \texttimes & 11.48 & 6.39 & 12.65 & 18.44 & 11.20 & 11.94 & 5.93 & 12.32 & 15.43 & 8.94 & 10.38 & 6.47 & 11.65 & 19.09 & 9.06 \\
    NSA \cite{baltatzis2024nsa} & \texttimes & 7.83 & 7.33 & -- & -- & 13.12 & -- & -- & -- & -- & -- & -- & -- & -- & -- & -- \\
    S-MotionGPT$^*$ \cite{jiang2023motiongpt} & \texttimes & 11.23 & 4.39 & 12.41 & 13.74 & 11.45 & 10.81 & 3.78 & 11.58 & 11.31 & 8.82 & 9.45 & 3.41 & 10.42 & 9.08 & 9.68 \\
    
    \midrule
    SOKE (ours) & \checkmark & \textbf{6.82} & \textbf{2.35} & \textbf{7.75} & \textbf{10.08} & \textbf{14.48} & \textbf{6.24} & \textbf{1.71} & \textbf{7.38} & \textbf{9.68} & \textbf{11.30} & \textbf{4.77} & \textbf{1.38} & \textbf{6.04} & \textbf{7.72} & \textbf{11.87}
 \\
    
    \bottomrule
    \end{tabular}
    }
    \vspace{-2mm}
    \caption{Comparison with state-of-the-art sign language generation (text-to-sign) methods. $^*$denotes reimplementation results.}
    \label{tab:sota_slg}
    \vspace{-2mm}
\end{table*}

\begin{table*}[t]
\small
\setlength\tabcolsep{5pt}
    \centering
    \begin{tabular}{l|c|cccc|cccc|cccc}
    \toprule
    \multirow{3}{*}{\shortstack{Decoding\\Method}} & \multirow{3}{*}{\shortstack{Sign\\Retrieval}} & \multicolumn{4}{c|}{How2Sign} & \multicolumn{4}{c|}{CSL-Daily} & \multicolumn{4}{c}{Phoenix-2014T} \\
    & & \multicolumn{3}{c}{DTW-PA-JPE$\downarrow$} & Latency$\downarrow$ & \multicolumn{3}{c}{DTW-PA-JPE$\downarrow$} & Latency$\downarrow$ & \multicolumn{3}{c}{DTW-PA-JPE$\downarrow$} & Latency$\downarrow$ \\
    \cmidrule(r){3-5}\cmidrule(r){6-6}\cmidrule(r){7-9}\cmidrule(r){10-10}\cmidrule(r){11-13}\cmidrule(r){14-14}
    & & Avg & Body & Hand & s/video & Avg & Body & Hand & s/video & Avg & Body & Hand & s/video \\
    
    \midrule
    Sequential & \texttimes & 4.64 & 8.55 & 3.53 & 3.26 & 3.74 & 8.14 & 2.49 & 3.15 & 3.26 & 6.97 & 2.20 & 3.23 \\
    Parallel & \texttimes & 5.06 & 9.27 & 3.86 & 1.39 & 4.19 & 8.61 & 2.93 & 1.41 & 3.54 & 7.02 & 2.55 & 1.35 \\
    Multi-head & \texttimes & 4.17 & 7.91 & 3.10 & 1.46 & 3.37 & 7.58 & 2.17 & 1.45 & 2.80 & 6.16 & 1.85 & 1.44 \\
    Multi-head & \checkmark & \textbf{3.34} & \textbf{6.82} & \textbf{2.35} & 1.55 & \textbf{2.72} & \textbf{6.24} & \textbf{1.71} & 1.52 & \textbf{2.13} & \textbf{4.77} & \textbf{1.38} & 1.51 \\
    
    \bottomrule
    \end{tabular}
    \vspace{-2mm}
    \caption{Ablation study for the decoding method and sign retrieval. Latencies are measured using a single RTX 3090 GPU.}
    \label{tab:abl_dec_rag}
    \vspace{-4mm}
\end{table*}

\subsection{Comparison with State-of-the-Art Methods}
\noindent\textbf{Qualitative Comparison.} 
We first reimplement a state-of-the-art (SOTA) open-sourced motion generation approach, MotionGPT \cite{jiang2023motiongpt}, which utilizes a VQ-VAE to discretize whole-body motions and employs a monolingual language model \cite{t5}, on sign language datasets. The resulting model, S-MotionGPT, achieves competitive results on sign language benchmarks (Table \ref{tab:sota_slg}). In Figure \ref{fig:vis}, we perform a qualitative comparison between our SOKE and S-MotionGPT. As can be easily seen, our proposed method produces significantly better results, with visual quality comparable to the ground truth sign videos across the ASL, CSL, and DGS datasets. Notably, our method can generate precise and informative hand movements, which play a key role in conveying semantics in sign languages. Additionally, the proposed method supports three sign languages simultaneously, whereas S-MotionGPT is monolingual and requires separate models for each sign language dataset.

\noindent\textbf{Quantitative Comparison.}
In addition to the above qualitative comparison, we also perform a quantitative evaluation against SOTA SLG methods, as detailed in Table \ref{tab:sota_slg}. 
Our reimplemented S-MotionGPT achieves competitive results, with low DTW errors of 4.39 and 3.78 on How2Sign and CSL-Daily, respectively, demonstrating the effectiveness of formulating SLG as a language modeling task.
The previous best SLG approach, Neural Sign Actors (NSA) \cite{baltatzis2024nsa}, formulates SLG as a motion generation task and employs a diffusion model as its backbone. However, this approach overlooks the linguistic properties of sign languages, resulting in higher average DTW errors compared to S-MotionGPT.
In contrast, our method sets a new state-of-the-art across all three benchmarks, achieving significantly lower DTW errors of 2.35, 1.71, and 1.38 on How2Sign, CSL-Daily, and Phoenix-2014T, respectively, using a single unified model.

\subsection{Ablation Study}
\begin{figure}[t]
\centering
\includegraphics[width=1.0\linewidth]{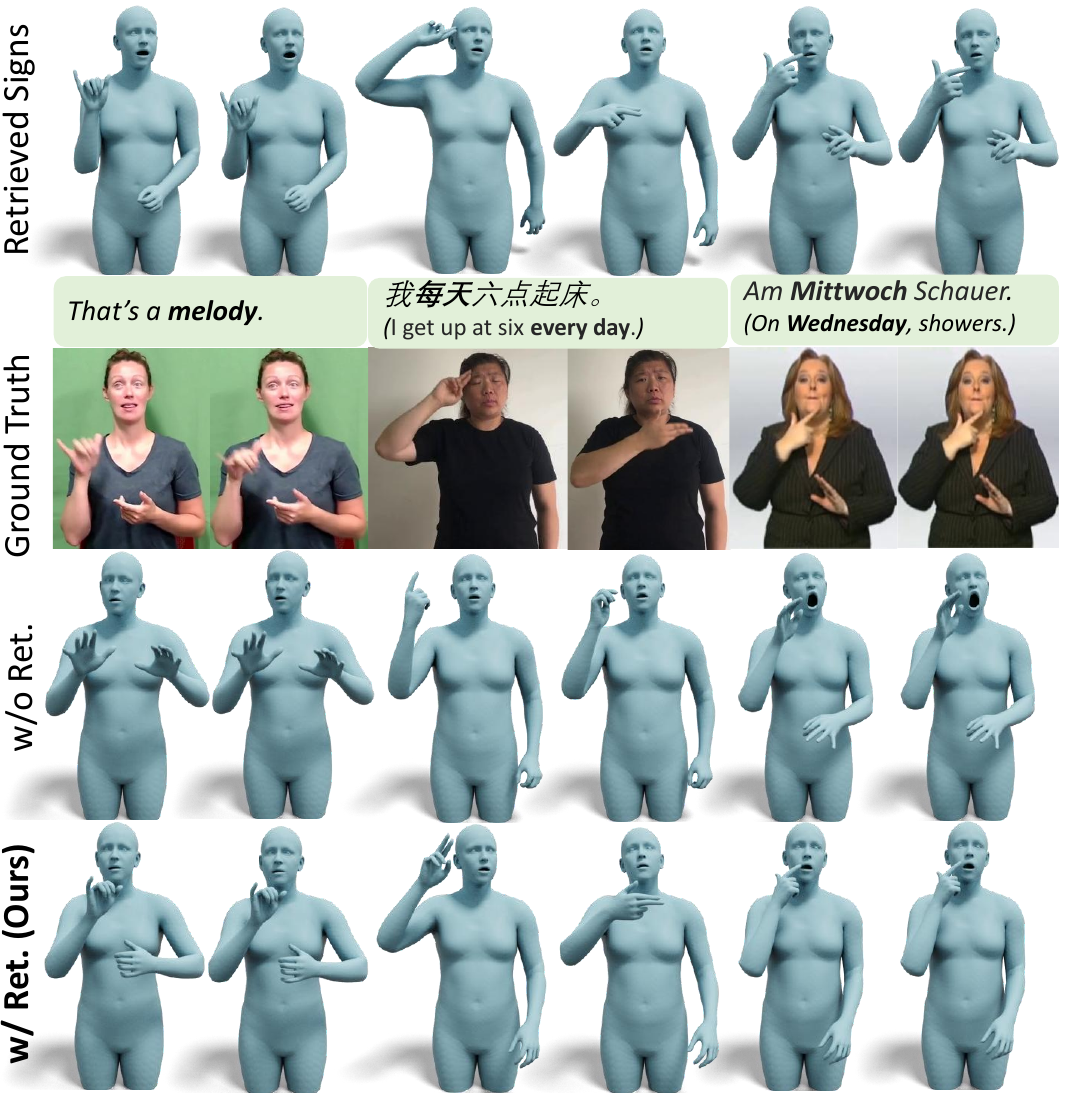}
\vspace{-7mm}
\caption{Qualitative ablation study for retrieval-enhanced SLG. (Left: How2Sign; Middle: CSL-Daily; Right: Phoenix-2014T.)}
\vspace{-5mm}
\label{fig:vis_rag}
\end{figure}

\begin{table*}[t]
\small
    \centering
    \begin{tabular}{ccc|cccccc|cccccc}
    \toprule
    \multirow{3}{*}{H2S} & \multirow{3}{*}{CSL} & \multirow{3}{*}{Ph-T} & \multicolumn{6}{c|}{Generation} & \multicolumn{6}{c}{Reconstruction} \\
    & & & \multicolumn{2}{c}{H2S (DTW$\downarrow$)} & \multicolumn{2}{c}{CSL (DTW$\downarrow$)} & \multicolumn{2}{c|}{Ph-T (DTW$\downarrow$)} & \multicolumn{2}{c}{H2S (JPE$\downarrow$)} & \multicolumn{2}{c}{CSL (JPE$\downarrow$)} & \multicolumn{2}{c}{Ph-T (JPE$\downarrow$)} \\
    \cmidrule(r){4-5}\cmidrule(r){6-7}\cmidrule(r){8-9}\cmidrule(r){10-11}\cmidrule(r){12-13}\cmidrule(r){14-15}
    & & & Body & Hand & Body & Hand & Body & Hand & Body & Hand & Body & Hand & Body & Hand \\
    \midrule
    
    \checkmark & & & 7.92 & 3.07 & -- & -- & -- & -- & 21.52 & 7.74 & -- & -- & -- & -- \\
    \checkmark & \checkmark &  & 7.11 & 2.63 & 6.79 & 2.14 & -- & -- & 19.88 & 6.97 & 24.29 & 5.83 & -- & -- \\   
    \checkmark & \checkmark & \checkmark & \textbf{6.82} & \textbf{2.35} & \textbf{6.24} & \textbf{1.71} & \textbf{4.77} & \textbf{1.38} & \textbf{19.37} & \textbf{6.65} & \textbf{23.52} & \textbf{5.13} & \textbf{25.79} & \textbf{6.78} \\
    
    \bottomrule
    \end{tabular}
    \vspace{-3mm}
    \caption{Study on the scalability of our approach. We use procrustes-aligned mean per joint position error (PA-MPJPE) \cite{lin2023osx, yu2025signavatars} to assess reconstruction performance of the decoupled tokenizer.  (Ph-T: Phoenix-2014T.)}
    \label{tab:abl_data}
    \vspace{-5mm}
\end{table*}

\begin{table}[t]
\small
\setlength\tabcolsep{3pt}
    \centering
    \resizebox{\linewidth}{!}{
    \begin{tabular}{ccc|cc|cc|cc}
    \toprule
    \multirow{2}{*}{VAE-1} & \multirow{2}{*}{VAE-2} & \multirow{2}{*}{VAE-3} & \multicolumn{2}{c|}{H2S (DTW$\downarrow$)} & \multicolumn{2}{c|}{CSL (DTW$\downarrow$)} & \multicolumn{2}{c}{Ph-T (DTW$\downarrow$)} \\
    \cmidrule(r){4-5}\cmidrule(r){6-7}\cmidrule(r){8-9}
    & & & Body & Hand & Body & Hand & Body & Hand \\
    \midrule
    
    Whole-body & \texttimes & \texttimes & 9.49 & 3.56 & 9.22 & 3.35 & 8.03 & 3.23 \\
    Body & Hands & \texttimes & 7.99 & 3.13 & 6.93 & 2.85 & 6.24 & 2.18 \\
    Body & L-hand & R-hand & \textbf{6.82} & \textbf{2.35} & \textbf{6.24} & \textbf{1.71} & \textbf{4.77} & \textbf{1.38} \\   
    
    \bottomrule
    \end{tabular}
    }
    \vspace{-3mm}
    \caption{Ablation study on the decoupled tokenizer.}
    \label{tab:abl_vae}
    \vspace{-3mm}
\end{table}

\noindent\textbf{Decoding Method.}
We compare the performance of different decoding methods in Table \ref{tab:abl_dec_rag}. The traditional sequential decoding method is widely used in existing motion generation approaches \cite{lu2024humantomato,yi2023generating}. However, since tokens from different body parts are flattened into a single sequence, the number of decoding steps is tripled, leading to high inference latency (3.26s per video on How2Sign).
By using three separate LM decoders, parallel decoding predicts token sequences for each body part simultaneously. This formulation significantly reduces latency (1.39s per video on How2Sign) but results in higher DTW errors due to the lack of information fusion across part-wise motion tokens.
Notably, our multi-head decoding method achieves a better trade-off between efficiency and information integration, maintaining comparable latency to parallel decoding while delivering even superior generation quality than sequential decoding.
We attribute this to the weighting mechanism (soft gating): since the decoder inputs are a weighted average of all part-wise tokens, errors in a single token may have a reduced impact on the overall performance.

\noindent\textbf{Sign Retrieval.}
By leveraging external sign dictionaries, we develop a retrieval-enhanced SLG approach that incorporates retrieved word-level signs as additional generation conditions. As shown in Table \ref{tab:abl_dec_rag}, this retrieval mechanism significantly enhances the precision of generated signs, reducing the average DTW errors by 19.9\%/19.6\%/23.9\% on How2Sign, CSL-Daily, and Phoenix-2014T, respectively.
Furthermore, we perform a qualitative comparison in Figure \ref{fig:vis_rag}. The figure demonstrates that the retrieved signs effectively guide the model's generation: similar pose patterns from the retrieved signs, particularly in hand shapes, appear in the generated outputs. This guidance substantially improves the precision of the generated signs.

\noindent\textbf{Scalability with Multilingual Data.}
Results in Table \ref{tab:abl_data} show that incorporating multilingual data significantly improves generation performance, demonstrating the scalability of our approach. This improvement stems from two key factors: 1) the tokenizer benefits from more pose training data (right half of Table \ref{tab:abl_data}), which subsequently enhances the generator's performance; 2) the presence of cross-lingual signs, as revealed in \cite{Wei_2023_ICCV}. For example, the sign for ``cold'' in CSL and DGS shares similar body movements. Training on multilingual data helps the model better understand and generalize across these cross-lingual signs.

\noindent\textbf{Decoupled Tokenizer.}
Inspired by the multi-cue characteristics of sign languages, we introduce a decoupled tokenizer that discretizes sign motion sequences for different body parts. In Table \ref{tab:abl_vae}, we present an ablation study on various configurations: a single VQ-VAE for whole-body motions and one VAE for the body and another for both hands. The findings indicate that progressively separating the body and hands leads to consistently improved performance.

\subsection{User Study}
In addition to the objective metrics presented in the previous tables, we conducted a user study involving 5 professional ASL signers and 4 professional CSL signers. These signers were asked to rate the alignment between generated signs and text annotations on a scale from 1 to 10, with higher scores indicating better semantic conveyance of the texts. Ground truth sign motions were also included in the study. Specifically, we provided 15 generated signs from both the baseline, S-MotionGPT, and our proposed method, SOKE. The order of the signs was randomly shuffled to prevent potential bias.

\begin{figure}[t]
\centering
\includegraphics[width=1.0\linewidth]{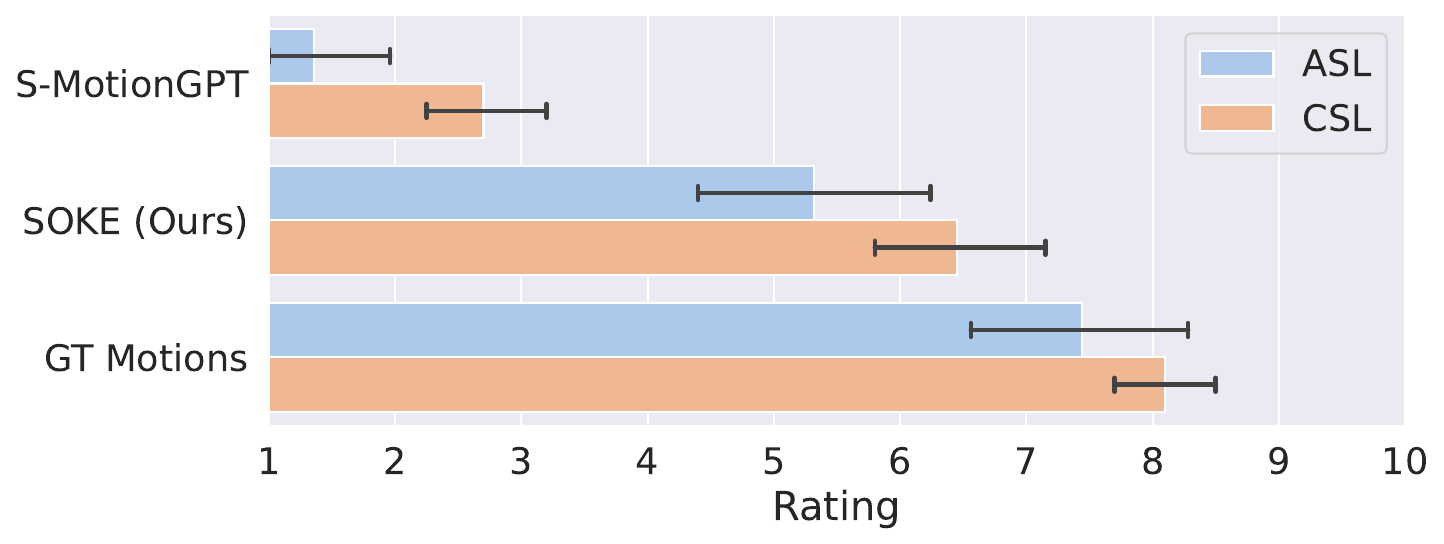}
\vspace{-7mm}
\caption{User study with professional ASL and CSL signers. We report average ratings of S-MotionGPT, our proposed SOKE, and ground truth motions.}
\vspace{-4mm}
\label{fig:user}
\end{figure}

As shown in Figure \ref{fig:user}, our method achieved an encouraging average rating of 5.32/6.45, significantly surpassing the baseline, which received a low score of 1.36/2.70 on How2Sign and CSL-Daily, respectively. 
These results align with the above qualitative and quantitative evaluations, indicating the potential of our method in constituting a two-way communication system between the deaf and hearing.

\vspace{-1mm}
\section{Conclusion}
\vspace{-1mm}
In this paper, we propose SOKE, a retrieval-enhanced multilingual sign language generator. Unlike existing methods that view sign language generation as a visual content generation task, we emphasize its linguistic nature, characterized by discrete semantic units. To capture the multi-cue characteristics of sign languages, we employ a decoupled tokenizer to map continuous motions into discrete tokens for the upper body and both hands. To balance efficiency and generation quality, we propose a novel multi-head decoding method that reduces decoding steps by two-thirds while dynamically integrating information from different body parts. Additionally, we present a simple yet effective retrieval-enhanced approach by leveraging external sign dictionaries. Extensive evaluations on a combined ASL-CSL-DGS dataset validate the effectiveness of our method.

\noindent\textbf{Acknowledgements.}
S. Zafeiriou and part of the research was funded by the
EPSRC Fellowship DEFORM (EP/S010203/1), EPSRC Project GNOMON
(EP/X011364/1) and Turing AI Fellowship (EP/Z534699/1). R.A. Potamias and R. Zuo were supported by EPSRC Project GNOMON (EP/X011364/1). This research was also supported by the Sovereign AI AIRR programme. We would also like to thank anonymous signers who participated in the user study.

\small
\bibliographystyle{ieeenat_fullname}
\bibliography{main}

\clearpage

\appendix
\renewcommand\thetable{S\arabic{table}}
\renewcommand\thefigure{S\arabic{figure}}
\renewcommand{\theequation}{S\arabic{equation}}

\twocolumn[{
\maketitle
\begin{center}
\includegraphics[width=1.0\textwidth]{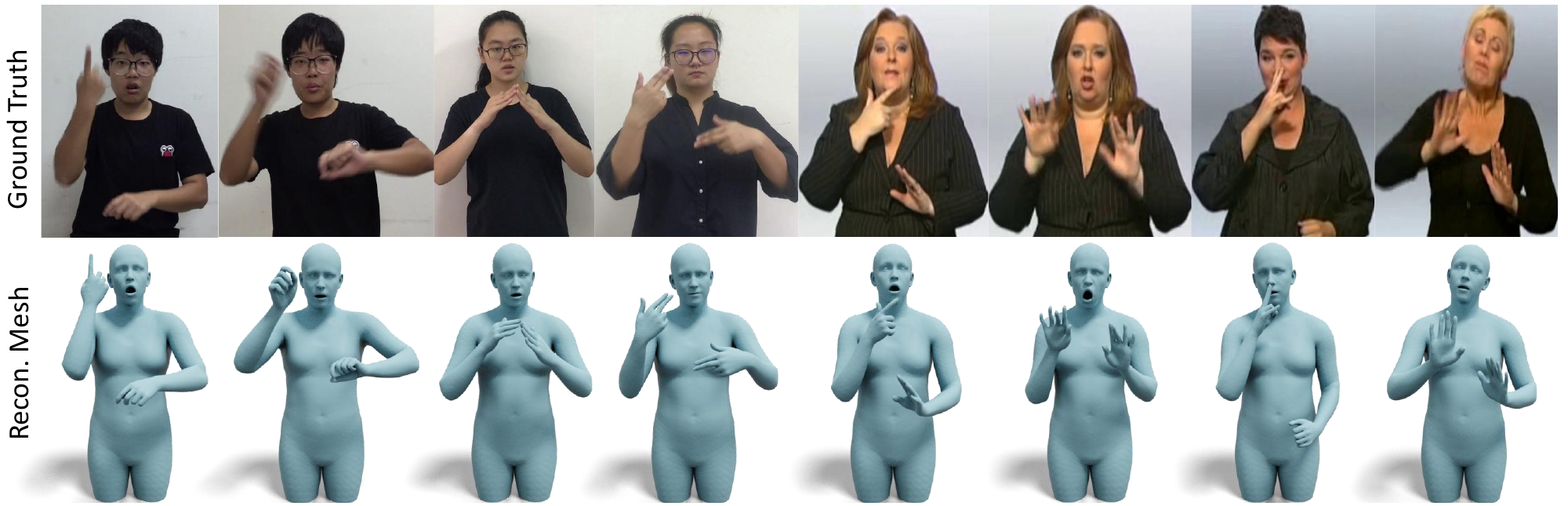}
\captionof{figure}{Qualitative comparisons between ground truth video
frames and reconstructed meshes obtained from the proposed
SMPL-X pose fitting pipeline on CSL-Daily (left) and Phoenix-2014T (right). Zoom-in for hand details.}
\label{fig:qual_fit}
\end{center}
}]

\section{Curating SMPL-X Poses}
To curate a high-fidelity dataset with accurate 3D annotations, we rely on state-of-the-art performing methods for 3D hand~\cite{potamias2024wilor} and body reconstruction~\cite{lin2023osx}. 
Specifically, given a 2D video of a signer, we first detect the number of identities in the video using an off-the-shelf detector~\cite{mediapipe} and retain the most confident detection box. 
Following that we feed the tight human crop to OSX~\cite{lin2023osx} to extract a rough human body pose estimation. 
Given that OSX often fails to accurately capture the arm positions and the hand poses, we follow a two-step approach that accurately refines the human pose. 
To accurately reconstruct the fine details of the hand poses, we utilize WiLoR~\cite{potamias2024wilor}, a state-of-the-art 3D reconstruction pipeline that can detect and reconstruct challenging hand poses with high fidelity. 
We acquire the hand poses of WiLoR along with the global orientation of the hand and directly substitute the hand parameters derived from OSX. 
In the second state, we employ Mediapipe body pose estimation~\cite{mediapipe} to extract 2D joint location $\mathbf{J}^{2D}$ for the shoulders and the arms. 
Using the derived joint locations, we employ an optimization scheme that refines the OSX poses of the upper body, while keeping the hand poses and orientation fixed:
\begin{equation}
    \mathcal{L}_{rec} = ||\mathbf{J}^{2D} - \Pi_K(\mathbf{\hat{J}}^{3D})||_1,
\end{equation}
where $\mathbf{\hat{J}}^{3D}$ are the predicted 3D joints and $\Pi_K$ is the weak-perspective projection. 
To further constrain the temporal coherence of the reconstructions, we include an additional temporal loss $\mathcal{L}_{temp}$: 
\begin{equation}
    \mathcal{L}_{temp} = || \mathbf{X}_{f} - \mathbf{X}_{f-1}||_2 + || \mathbf{J}_{f} - \mathbf{J}_{f-1}||_2,
\end{equation}
where $\mathbf{X}_f$ denotes the 3D mesh in frame $f$.
Finally, to penalize irregular poses, we include a pose regularization: 
\begin{equation}
    \mathcal{L}_{reg} = ||\theta||_2 
\end{equation}
that constrains irregular upper body poses. 

\begin{table}[t]
\small
    \centering
    \begin{tabular}{l|ccc}
    \toprule
    Method & Body$\downarrow$ & Left Hand$\downarrow$ & Right Hand$\downarrow$ \\
    \midrule

    FrankMoCap \cite{rong2021frankmocap} & 78.07 & 20.47 & 19.62\\
    PIXIE \cite{pixie} & 60.11 & 25.02 & 22.42\\
    PyMAF-X \cite{zhang2023pymaf} & 68.61 & 21.46 & 19.19\\
    SMPLify-X \cite{smplx} & 56.07 & 22.23 & 18.83\\
    SGNify \cite{forte2023reconstructing} & 55.63 & 19.22 & 17.50\\
    OSX \cite{lin2023osx} & 47.32 & 18.34 & 18.12\\
    NSA \cite{baltatzis2024nsa} & \textbf{46.42} & \underline{16.17} & \underline{15.23}\\
    
    \midrule
    Ours & \underline{46.73} & \textbf{10.55} & \textbf{8.94} \\
    
    \bottomrule
    \end{tabular}
    \caption{Reconstruction errors on SGNify mocap dataset \cite{forte2023reconstructing}. We report mean per vertex errors in mm.}
    \label{tab:fitting}
\end{table}

\begin{figure*}[t]
    \centering
    \includegraphics[width=\textwidth]{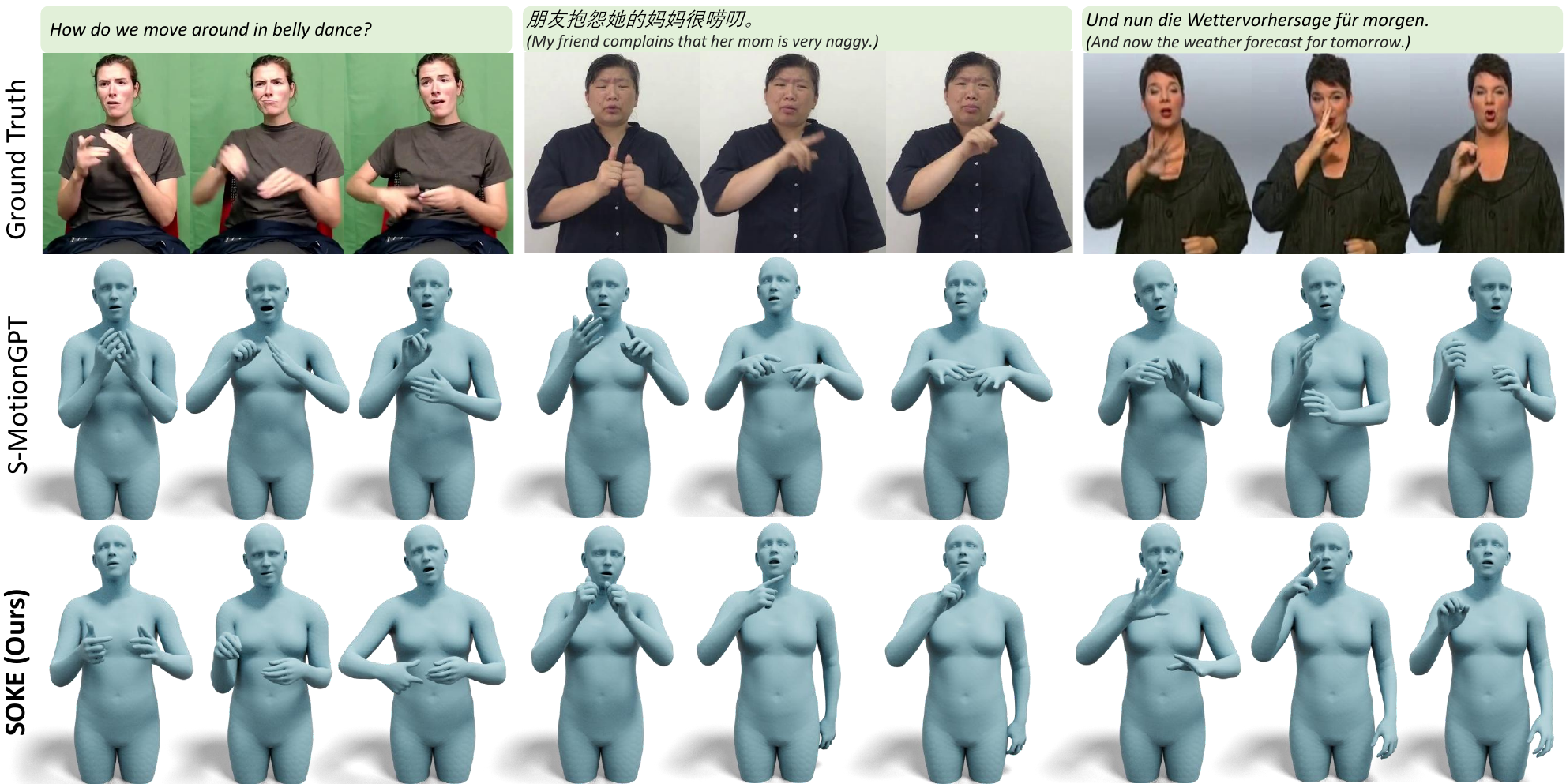}
    \caption{Qualitative comparisons of generated signs between our proposed method, SOKE, with the SOTA method, S-MotionGPT \cite{jiang2023motiongpt}, on the test sets of How2Sign (left), CSL-Daily (middle), and Phoenix-2014T (right).}
    \label{fig:supp_gen}
\end{figure*}

Since neither CSL-Daily \cite{csl-daily} nor Phoenix-2014T \cite{2014T} provides 3D annotations, we perform qualitative evaluations, as illustrated in Figure \ref{fig:qual_fit}. The results clearly demonstrate that the proposed pose fitting pipeline can accurately reconstruct 3D hands and is robust across various handshapes. To quantitatively assess the pipeline, we further apply it to the SGNify mocap dataset \cite{forte2023reconstructing}, which includes 57 signs with annotated meshes. The results presented in Table \ref{tab:fitting} indicate that our method achieves the lowest hand reconstruction errors and comparable body errors to the previous best method \cite{baltatzis2024nsa}, establishing our approach as a powerful tool for curating more sign language datasets in the future.

\section{Additional Qualitative Results}
\begin{figure}[t]
\centering
\includegraphics[width=1.0\linewidth]{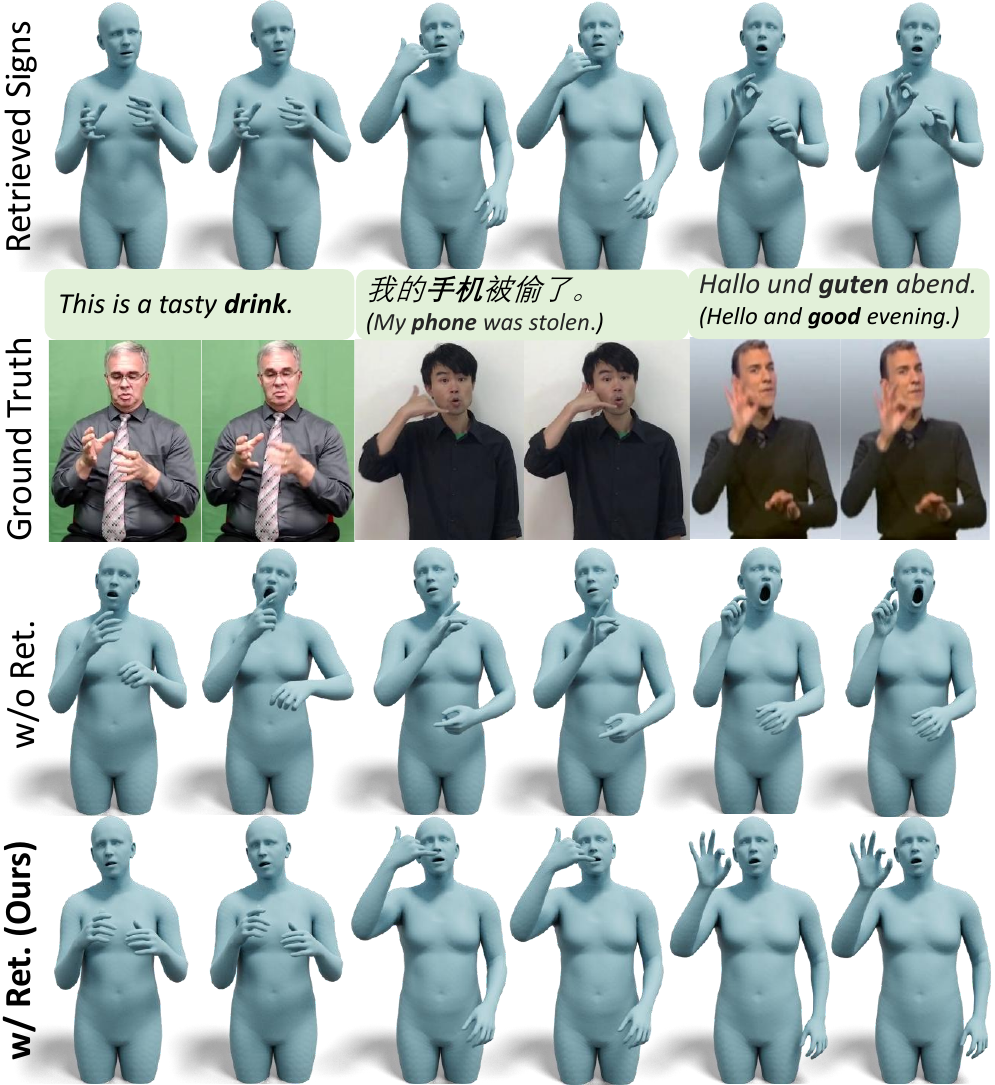}
\caption{Qualitative ablation study for retrieval-enhanced SLG. (Left: How2Sign; Middle: CSL-Daily; Right: Phoenix-2014T.)}
\label{fig:supp_ret}
\end{figure}

Please refer to our project page for video demonstrations of generated signs. These demos include ground truth sign videos, as well as generations from the SOTA method, S-MotionGPT \cite{jiang2023motiongpt}, and our proposed SOKE. 
Additionally, we provide several qualitative results to showcase the generated signs (Figure \ref{fig:supp_gen}) and highlight the effectiveness of our retrieval-enhanced SLG approach (Figure \ref{fig:supp_ret}).

\section{Additional Quantitative Results}
\begin{table}[t]
\small
\setlength\tabcolsep{3pt}
    \centering
    \begin{tabular}{cc|cc|cc|cc}
    \toprule
    \multirow{2}{*}{$N_Z^B$} & \multirow{2}{*}{$N_Z^{LH}=N_Z^{RH}$} & \multicolumn{2}{c|}{H2S (JPE$\downarrow$)} & \multicolumn{2}{c|}{CSL (JPE$\downarrow$)} & \multicolumn{2}{c}{Ph-T (JPE$\downarrow$)}  \\
    \cmidrule(r){3-4}\cmidrule(r){5-6}\cmidrule(r){7-8}
    & & Body & Hand & Body & Hand & Body & Hand \\
    
    \midrule
    96 & 128 & 19.37 & 7.07 & 23.52 & 5.80 & 25.79 & 7.35 \\
    96 & 256 & 19.37 & 6.86 & 23.52 & 5.52 & 25.79 & 7.11 \\
    64 & 192 & 20.04 & 6.65 & 24.13 & 5.13 & 26.02 & 6.78 \\
    128 & 192 & 19.95 & 6.65 & 23.91 & 5.13 & 26.27 & 6.78 \\
    96 & 192 & \textbf{19.37} & \textbf{6.65} & \textbf{23.52} & \textbf{5.13} & \textbf{25.79} & \textbf{6.78} \\
    
    \bottomrule
    \end{tabular}
    \caption{Study on the codebook sizes for the body ($N_Z^B$) and hands ($N_Z^{LH}, \ N_Z^{RH}$). We use procrustes-aligned mean per joint position error (PA-MPJPE) to assess the reconstruction performance of the decoupled tokenizer.}
    \label{tab:cb_size}
\end{table}

\begin{table}[t]
\small
    \centering
    \begin{tabular}{c|cc|cc|cc}
    \toprule
    \multirow{2}{*}{$\lambda$} & \multicolumn{2}{c|}{H2S (DTW$\downarrow$)} & \multicolumn{2}{c|}{CSL (DTW$\downarrow$)} & \multicolumn{2}{c}{Ph-T (DTW$\downarrow$)}  \\
    \cmidrule(r){2-3}\cmidrule(r){4-5}\cmidrule(r){6-7}
    & Body & Hand & Body & Hand & Body & Hand \\
    
    \midrule
    0.1 & 7.95 & 2.82 & 7.46 & 2.13 & 5.47 & 2.04 \\
    0.2 & 7.28 & 2.76 & 6.91 & 1.95 & 5.08 & 1.68 \\
    1/3 & \textbf{6.82} & \textbf{2.35} & \textbf{6.24} & \textbf{1.71} & \textbf{4.77} & \textbf{1.38} \\
    0.4 & 7.34 & 2.62 & 7.11 & 1.91 & 6.39 & 1.96 \\
    
    \bottomrule
    \end{tabular}
    \caption{Study on the impact of $\lambda$, a hyper-parameter used for fusing part-wise token embeddings in our multi-head decoding method.}
    \label{tab:lambda} 
\end{table}

\noindent\textbf{Codebook Size.}
We perform a hyper-parameter analysis on the codebook sizes for the body ($N_Z^B$) and hands ($N_Z^{LH}, \ N_Z^{RH}$) in our decoupled tokenizer. As shown in Table \ref{tab:cb_size}, we find that using either larger or smaller codebooks results in degraded reconstruction performance. Our default configuration ($N_Z^B=96$, $N_Z^{LH}=N_Z^{RH}=192$) delivers the best performance among all settings.

\noindent\textbf{Impact of $\lambda$ on SLG.}
In our multi-head decoding method, we introduce a hyper-parameter, $\lambda$, to control the weight of hand tokens during embedding fusion. The results in Table \ref{tab:lambda} demonstrate that $\lambda = 1/3$, \ie, assigning equal weights to the body and hands, yields the best performance. This further underscores the importance of each body part in conveying the semantics of sign languages.

\begin{table}[t]
\small
\setlength\tabcolsep{2pt}
\centering
\resizebox{1.0\linewidth}{!}{
\scriptsize
\begin{tabular}{l|c|ccc|ccc|ccc}
    \toprule
    \multirow{2}{*}{Method} & \multirow{2}{*}{\shortstack{Multi\\ling.}} & \multicolumn{3}{c|}{H2S (DTW$\downarrow$)} & \multicolumn{3}{c|}{CSL (DTW$\downarrow$)} & \multicolumn{3}{c}{Phoenix (DTW$\downarrow$)} \\
    & & Avg & Body & Hand & Avg & Body & Hand & Avg & Body & Hand \\
    \midrule
    S-MotionGPT & \texttimes & 5.91 & 11.23 & 4.39 & 5.34 & 10.81 & 3.78 & 4.75 & 9.45 & 3.41 \\ 
    Ours & \texttimes & 4.14 & 7.92 & 3.07 & 4.18 & 8.18 & 3.04 & 3.83 & 7.25 & 2.85 \\
    Ours & \checkmark & \textbf{3.34} & \textbf{6.82} & \textbf{2.35} & \textbf{2.72} & \textbf{6.24} & \textbf{1.71} & \textbf{2.13} & \textbf{4.77} & \textbf{1.38}  \\
    \bottomrule
\end{tabular}
}
\caption{Performance of our method on monolingual datasets.}
\label{tab:single}
\end{table}

\noindent\textbf{Monolingual Performance.} 
As shown in Table \ref{tab:single}, our method still outperforms the SOTA method, S-MotionGPT, when training on monolingual SL datasets, while the best results are achieved by the multilingual version of our method.

\section{Illustration of Decoupled Tokenizer}
\begin{figure}[t]
\centering
\includegraphics[width=1.0\linewidth]{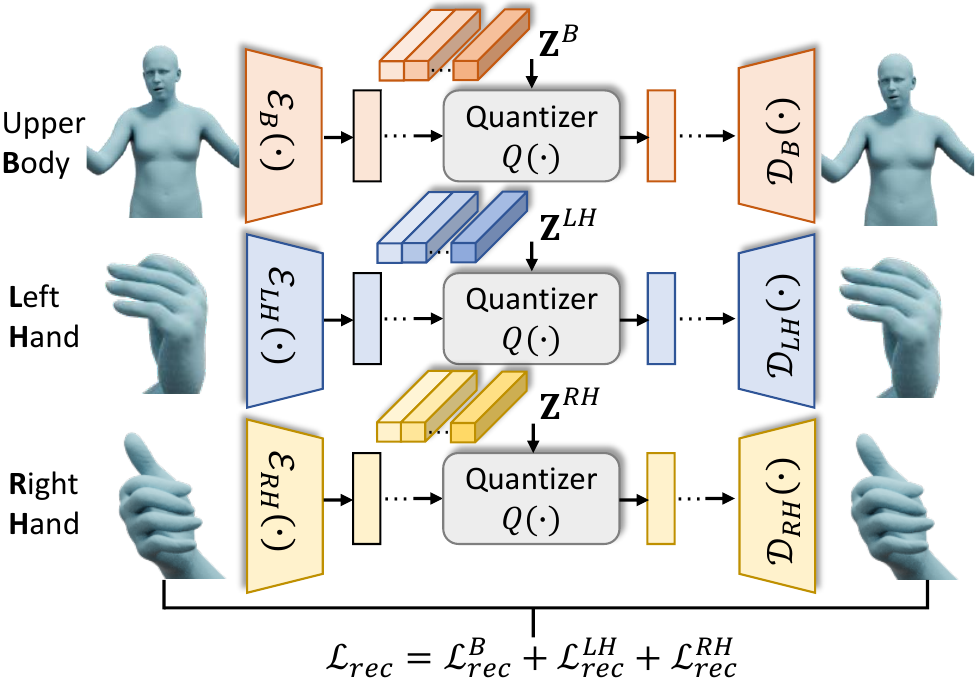}
\caption{Workflow of our decoupled tokenizer. It is composed of three parallel VQ-VAEs, each dedicated to generating motion tokens for a different part of the signer's body: the upper body, left hand, and right hand. }
\label{fig:tokenizer}
\end{figure}

As shown in Figure \ref{fig:tokenizer}, we provide an illustration of our decoupled tokenizer for better understanding. It utilizes three VQ-VAEs to model the key regions of a signer: the upper body, left hand, and right hand.

\section{Discussion}
\noindent\textbf{Broader Impacts.}
Sign language is the primary mode of communication for the deaf communities. Due to significant grammatical differences from spoken languages, a notable communication gap exists between the deaf and hearing individuals. In this work, we propose an autoregressive sign language model, which is capable of generating multilingual sign language avatars from text inputs within a single unified framework. Extensive quantitative and qualitative results suggest the potential of our method to form a practical deaf-hearing communication system.

\noindent\textbf{Limitations.}
Our method employs 3D avatars to represent signers, enabling high-fidelity motion representations. However, there is a lack of 3D annotations in existing sign language datasets. While our proposed SMPL-X pose fitting pipeline can accurately reconstruct 3D meshes from 2D keypoints, some reconstruction errors are inevitable. In the future, the release of more sign language datasets with annotated meshes is anticipated, which could significantly enhance avatar-based sign language generation models.

\noindent\textbf{Future Works.}
We have validated the proposed multilingual sign language generator on three widely-adopted sign languages, Chinese, American, and German sign language \cite{csl-daily,duarte2021how2sign,2014}. As the scalability of our approach has been demonstrated in Table 3 of the main paper, in the future, we plan to extend our method to support more sign languages, such as British Sign Language \cite{albanie2020bsl} and Indian Sign Language \cite{joshi2024isign}.

\end{document}